\documentclass[conference]{IEEEtran}
\IEEEoverridecommandlockouts
\usepackage{xr-hyper}
\usepackage{hyperref}
\usepackage{cite}
\usepackage{amsmath,amssymb,amsfonts}
\usepackage{algorithmic}
\usepackage{graphicx}
\usepackage{textcomp}
\usepackage{xcolor}
\usepackage{url}
\usepackage{listings}
\usepackage{xspace}
\usepackage{cuted}
\usepackage{stfloats}
\usepackage{comment}
\usepackage{subcaption}
\usepackage{multirow}
\usepackage{threeparttable}
\usepackage{array}
\usepackage{longtable}
\usepackage{makecell}
\usepackage{booktabs}
\usepackage{soulpos}
\usepackage{xcolor}

\ulposdef{\hlst}{%
    \rlap{\textcolor{red}{\rule[-.75ex]{\ulwidth}{2.5ex}}}%
    \rule[.45ex]{\ulwidth}{.1ex}%
}

\newcommand{\deletetext}[1]{\xspace}
\newcommand{\edittext}[2]{#2}
\newcommand{\newtext}[1]{#1}

\newcommand{\sts}{Seq2Seq\xspace}

\newcommand{\pmn}{PMN\xspace}
\newcommand{\dialogpt}{DialoGPT\xspace}
\newcommand{\pe}{\texttt{PolyEncoder}\xspace}
\newcommand{\colbert}{\texttt{ColBERT}\xspace}

\newcommand{\fancypcpe}{\ul{P}ersona-\ul{C}oded \ul{P}oly-\ul{E}ncoder\xspace}
\newcommand{\pcpe}{\texttt{PCPE}\xspace}
\newcommand{\pcpetext}{\texttt{PCPE-TEXT}\xspace}
\newcommand{\pcpekv}{\texttt{PCPE-KV}\xspace}

\newcommand{\etal}{\textit{et al.}\xspace}

\newcommand{\tfidf}{\texttt{TF-IDF}\xspace}

\newcommand{\selfattention}{\mbox{\texttt{S-Attn}}\xspace}
\newcommand{\multiattention}{\mbox{\texttt{M-Attn}}\xspace}
\newcommand{\colbertfusion}{\texttt{Col-Fuse}\xspace}

\newcommand{\real}{\mathbb{R}\xspace}

\newcommand{\softmax}{\texttt{softmax}\xspace}

\newcommand{\identity}[1]{\mathbb{I}({#1})\xspace}

\newcommand{\hr}{\texttt{HR}\xspace}
\newcommand{\hrk}[1]{\texttt{HR@{#1}}\xspace}
\newcommand{\mrr}{\texttt{MRR}\xspace}

\newcommand{\fscore}{\texttt{F1}\xspace}
\newcommand{\bleu}{\texttt{BLEU}\xspace}
\newcommand{\bleun}[1]{\texttt{BLEU{#1}}\xspace}

\newcommand{\stream}{\mathcal{S}\xspace}
\newcommand{\PEstream}{$\mathcal{S}_{\texttt{PE}}$\xspace}
\newcommand{\PCstream}{$\mathcal{S}_{\texttt{PC}}$\xspace}
\newcommand{\fuse}{\rho\xspace}

\def\BibTeX{{\rm B\kern-.05em{\sc i\kern-.025em b}\kern-.08em
   T\kern-.1667em\lower.7ex\hbox{E}\kern-.125emX}}
    
\begin{document}

\title{Persona-Coded Poly-Encoder: Persona-Guided Multi-Stream Conversational Sentence Scoring}

\author{
\IEEEauthorblockN{
Junfeng Liu\IEEEauthorrefmark{1}\IEEEauthorrefmark{3},
Christopher Symons\IEEEauthorrefmark{2}, and
Ranga Raju Vatsavai\IEEEauthorrefmark{2}\IEEEauthorrefmark{3}}
\IEEEauthorblockA{
\IEEEauthorrefmark{1}
\textit{Lirio AI Research, Lirio LLC}, 
Knoxville, TN, USA\\
\IEEEauthorrefmark{2}
\textit{Behavior Reinforcement Learning Lab, Lirio LLC}, 
Knoxville, TN, USA\\
\IEEEauthorrefmark{3}
\textit{Dept. of Computer Science, North Carolina State University}, 
Raleigh, NC, USA \\
\{jliu, csymons, rvatsavai\}@lirio.com
}
}

\IEEEoverridecommandlockouts
\IEEEpubid{\makebox[\columnwidth]{2375-0197/23/\$31.00~\copyright2023 IEEE \hfill} \hspace{\columnsep}\makebox[\columnwidth]{ }}

\IEEEaftertitletext{\vspace{-1\baselineskip}}

\maketitle

\begin{abstract} 
Recent advances in machine learning and deep learning have led to the widespread use of Conversational AI in many practical applications.
However, \edittext{leveraging}{it is still very challenging to leverage} auxiliary information that can provide conversational context or personalized tuning to improve the quality of conversations\deletetext{ is still very challenging}. For example, there has only been limited research on using an individual’s persona information to improve conversation quality, and even state-of-the-art conversational AI techniques are 
unable to effectively leverage signals from heterogeneous sources
of auxiliary data, such as multi-modal interaction data, demographics, SDOH data, etc.
In this paper, we present a novel Persona-Coded Poly-Encoder method that leverages
persona information in a multi-stream encoding scheme to improve the
quality of response generation for conversations. 
To show the efficacy of the proposed method, we evaluate our method on two different persona-based conversational datasets, and compared against two state-of-the-art methods.
Our experimental results and analysis demonstrate that our method can improve
conversation quality over the baseline method Poly-Encoder by 3.32\% and 2.94\% in terms of \mbox{\bleu} score and \mbox{\hrk{1}}, respectively. More significantly, our method offers a path to better utilization of
multi-modal data in conversational tasks.
Lastly, our study outlines several challenges and future research directions for advancing personalized conversational AI technology.

\end{abstract}

\begin{IEEEkeywords}
Conversational AI, Dialogue Systems, Persona, Personalization, Multi-Modal Data
\end{IEEEkeywords}

\section{Introduction}
\label{sec:intro}

Practical uses of conversational agents\deletetext{, such as chatbots,} have increased dramatically in recent years by leveraging
advances in
natural language processing\edittext{ and machine learning, especially}{, machine learning and} deep learning techniques.
Today, these agents play important roles in automated customer service, personal assistants, healthcare, and more. 
Many applications require \edittext{a conversational model}{an agent} to perform at a level that is
comparable to or surpasses human performance in terms of understanding the 
current state, personalizing responses, and other standards. 
In certain fields, such as healthcare, these requirements are particularly critical, 
as conversational topics can often be much more personal and sensitive.

Traditional conversational AI methods are often hard to scale due to 
strict requirements around data and supporting technologies, such as a
well-constructed knowledge graph or database, and excessive API calls for 
external dependencies or real-time information.
They also often require domain expertise and human intervention for evaluation. 
These requirements have largely limited the ability of traditional
\deletetext{conversational AI} methods to expand to many potentially valuable, but complex, use cases. 

Recent advances in deep learning have opened up tremendous opportunities
to expand the capabilities and scalability of conversational AI. 
Sequence-to-sequence (\sts) models~\cite{sutskever2014sequence} and transformers~\cite{vaswani2017attention, radford2019language, brown2020language}
are widely used to capture the basic characteristics of a conversation, such as
language flow and grammar. 
Beyond that, a significant amount of research effort around deep
conversational AI has attempted to leverage auxiliary resources that could
improve and personalize conversations. These auxiliary resources typically
supplement the conversational context with information beyond the language
itself, such as personas of the speakers~\cite{li2016persona, zhang2018personalizing}, 
the environments in which the speakers are interacting~\cite{mostafazadeh2017image}, 
external knowledge-bases~\cite{ghazvininejad2018knowledge}, etc.
These approaches also inspire and facilitate many applications in the industry. 
For example, one of the target applications that we are developing is a persona-driven conversational agent to improve patients’ experience and drive them towards better health outcomes. In addition to presenting a generic machine learning framework for combining multi-modal data, we also present several open challenges in building persona-guided conversational agents. 

In this paper: we (i) present a novel multi-stream \fancypcpe (\pcpe) network, (ii) provide the design and evaluation of various post-fusion strategies for auxiliary data, and (iii) experimentally evaluate the superiority of the proposed \pcpe network as compared to the state-of-the-art pre-fusion-based, single-stream networks.
Specifically, the \mbox{\pcpe} method outperforms the state-of-the-art \mbox{\colbert} method in response retrieval by 3.32\% (\bleu) and 2.94\% (\hrk{1}), respectively. Moreover, it presents a flexible framework that can better utilize multi-modal heterogeneous data via separate encoding streams.
This paper is organized as follows. 
Section~\ref{sec:related-work} reviews the current state-of-the-art
methods around persona-based conversational AI and their limitations.
In Section~\ref{sec:problem}, we define the problem and give an overview of single-stream pre-fusion and multi-stream post-fusion frameworks. 
Section~\ref{sec:method} presents our  method \pcpe that effectively utilizes persona data via a multi-stream post-fusion framework,
followed by an experimental comparison against two state-of-the-art methods 
in Section~\ref{sec:experiments} and~\ref{sec:exp:results}. We also highlight the 
advantages of \pcpe and observe challenges with existing datasets and evaluation
of personalization in Section~\ref{sec:discussion}, followed by conclusions in Section~\ref{sec:conclusion}.

\IEEEpubidadjcol

\section{Related Work}
\label{sec:related-work}

\subsection{Neural Conversational AI}
Neural approaches, particularly deep learning approaches, 
have attracted a lot of interest in conversational AI applications
due to their wide success in many fields of
natural language processing.
%
%
Depending on how the responses are generated, there are two main 
categories of neural conversational methods: generation-based and retrieval-based methods.

\textbf{Generation-based} methods~\cite{li2016persona, zhang2018generating, zhang2019dialogpt}
typically sample a novel sequence of words with a\deletetext{ \sts-based or transformer-based} decoder from a probabilistic distribution of all possible word tokens
conditioned on the inputs.
These novel responses are not limited by the existing data and can be used in
certain situations where creativity or novelty is desired.
%
%
Zhang~\etal~\cite{zhang2018generating} developed\deletetext{ a conversation method with} a
generative adversarial network to promote response diversity. 
%
Zhang~\etal trained \dialogpt~\cite{zhang2019dialogpt} \edittext{based on GPT-2~\mbox{\cite{radford2019language}} that encodes long-term dialogue history in the context.}{that encodes long-term dialogue history in the context with GPT-2~\mbox{\cite{radford2019language}}.}
\edittext{More recently, ChatGPT~\mbox{\cite{openai2022chatgpt}} was trained with 
reinforcement learning from human feedback and has attracted lots of attention
due to its ability to conduct conversations in various chat scenarios.
In practice, many generation-based applications suffer
from unstable quality of the responses, such as grammatical errors,
poor language flow, broken logic, and ignorance of facts.
Even though applications like ChatGPT have achieved great success, the exponentially 
growing sizes of such large language models (LLMs) make training very expensive.
}{In practice, generative methods often suffer
from unstable quality of the responses, such as poor language flow, broken logic, and ignorance of facts.
Large language models (LLMs) like ChatGPT~\mbox{\cite{openai2022chatgpt}} have achieved great success
in various chat scenarios. However, the exponentially 
growing sizes and hardware requirements of such LLMs made training very expensive.
}

\textbf{Retrieval-based} (or ranking-based) 
methods~\cite{zhang2018personalizing, wolf2019transfertransfo, humeau2019poly} select a 
response from an existing set of prescribed candidates, typically by learning the
similarities between the context and candidates through deep 
encoders and then scoring the candidates.
%
%
Humeau~\etal~\cite{humeau2019poly} proposed the \pe that is able to leverage high-level interactions between
the input query and candidates, and maintain computational efficiency through pre-calculation and caching. 
Although incapable of providing novel responses, 
retrieval models ensure the quality of the responses
because the candidates are careful crafted. 
They can also be easily extended to other applications by simply replacing
the candidate pool without changing the model, especially when certain 
responsive strategies are desired. 

Many document retrieval methods~\cite{luan2021sparse, khattab2020colbert} can 
also be easily engineered as retrieval-based solutions to dialogue systems.
Khattab~\etal~\cite{khattab2020colbert} proposed the \colbert method for effective
passage retrieval by using contextualized late interaction that leverages token-level 
similarity between the query and the candidate document.
However, these methods do not always fit conversation tasks due to the natural
differences between dialogues 
and long text documents (e.g., the 
lengths of the utterances/documents, the location of the key  ideas, the unique contexts for different tasks, etc).

\subsection{Persona-based Conversational AI}
\label{sec:related-work:persona}
In many real-world applications, personalization is desired or required 
for the dialogue system. 
For example, precision nudging~\cite{copper2020precision_nudging} applications
provide communications that drive patients to adopt certain
healthy behaviors, which require personalization to be effective across a diverse population. 
Efforts have been made towards leveraging personas into various applications including chit chat~\cite{li2016persona, zhang2018personalizing} and empathetic chat~\cite{rashkin2018towards}.
Personas have been shown to help gain user confidence and significantly improve the quality of conversations in recent \mbox{research~\cite{zhong2020towards, song2019exploiting}}.
Liu~\etal~\cite{liu2022persona} experimentally showed
that personas are critical to improve the dialogue performance
with two benchmark methods on the ConvAI2 Dataset~\cite{dinan2020second}. 
In this paper, we follow the definition of persona in~\mbox{\cite{liu2022persona}} as any type or format of data that contains personal information about a conversational partner that could help the model to understand the conversational context and provide better responses to the recipient.


One goal of personalized conversation is to address the speaker-consistency issue
when a model responds differently to the same query at two inferences because both query-response
pairs were seen during training, which confuses the model. 
%
Li~\etal~\cite{li2016persona} and Gu~\etal~\cite{gu2021deep}
leverage trainable speaker identity embeddings in existing \sts and BERT architectures,
respectively, to tackle this issue.
However, these methods fail to account for new speakers who never appeared in the data because
they rely only on the speaker ID during training.
Moreover, they are still unable to effectively provide personalized responses, partially due
to the lack of actual persona data in existing benchmark dialogue datasets, such as
Reddit~\cite{mazare2018training} or Twitter~\cite{ritter2010unsupervised} dataset.
%

Zhang~\etal~\cite{zhang2018personalizing} created the Persona-Chat dataset and made it possible to 
leverage actual persona information in dialogue systems. 
They encoded the textual persona entries of the speakers with profile memory networks into the dialogue
context for generation and retrieval tasks.
This dataset was further extended and used in the NeurIPS ConvAI2 
challenge~\cite{dinan2020second}. 
The winning method, TransferTransfo~\cite{wolf2019transfertransfo}, and the later \pe method~\cite{humeau2019poly} concatenate the persona and query before modeling in pre-fusion-based ways. 
These methods presented effective and innovative ways to explicitly encode the speaker profiles 
with long-term dialogue history to supplement the conversation context.
%
However, these datasets and methods did not address how to handle 
auxiliary data that might come in different formats or modalities.

Wang~\etal~\cite{wang2019persuasion} developed the PersuasionForGood dataset that consists of dialogues with persona data in the form of key-value (KV) attribute-based demographic features and psychological survey assessment scores. 
These KV attributes contain both categorical and numerical values.
Wang~\etal focused on a text classification problem to identify the persuasion strategy used in the conversations. 
Although they didn't propose solutions for conversation tasks, KV attributes are a common form of 
persona data and present challenges to existing pre-fusion methods since these personas cannot be directly concatenated with the textual inputs.

\section{Problem Definition}
\label{sec:problem}
In this paper, we consider a retrieval-based responses selection problem
formulated as follows: 
given an input triplet $x = (q, P, H)$ as the \newtext{multi-input} conversation context,
where 
$q$ is the text input query or utterance, 
$P=\{p_1, \cdots ,p_j\}$ is the set of persona entries (text- or attribute-based) and
$H = [h_1, \cdots ,h_k]$ is the list of dialogue histories, 
we want to train a model $f: (x, c_i) \rightarrow \mathbb{R}$ to 
assign a score $s_i$ to a candidate $c_i$ from a set of candidate 
responses $C$, then select a best response 
$c^* = \texttt{argmax}_{c_i\in C} f(x, c_i)$.
Here, we consider a scoring system that learns an embedder $g(x)$, which represents the conversation context into a $d$-dimension latent space such that $h = g(x) \in \real^d$,
and calculates the similarity with the candidate embedding $c_i \in \real^d$.
Examples of the \edittext{training data}{dataset} will be provided in Section~\mbox{\ref{sec:exp:dataset}}.

\deletetext{With multiple inputs $I_1, \cdots, I_n$ to the model, existing pre-fusion-based methods~\mbox{\cite{humeau2019poly, chen2020sequential}} concatenate them directly as one input and process it with a single embedder.
For example, Humeau~\mbox{\etal~\cite{humeau2019poly}} encoded the conversation context as $h = g([p_1; \cdots; p_j; h_1; \cdots; h_k; q])$. 
The pre-fusion methods implicitly assume that there is an ordering among the inputs, which
is not always true (e.g., the persona and query are independent). 
Moreover, these methods only work with homogeneous data (e.g., all inputs are textual) and would fail when inputs are heterogeneous, in terms of both the modality and the information they carry. 
}

\deletetext{A different approach is multi-stream embedding network (MSEN)~\mbox{\cite{gadiraju2020multimodal}} that
considers different inputs as 
heterogeneous and process them in different streams before fusion. That is,
$
h_1 = \stream_1(I_1), \hspace{5pt}
\cdots, \hspace{5pt}
h_n = \stream_n(I_n),
$
where $h_i \in \real^d$ is the independent embedding of the $i$-th input $I_i$ embedded by stream $\stream_i$.
Then the context embedding $h$ is produced as $h = \fuse([h1, \cdots, h_n])$,
where $\fuse(\cdot)$ is a fusion function such as a pooling layer
or an attention learning layer. 
Under MSEN, a model is able to embed multi-modal inputs with 
custom-engineered embedders, each of which is specialized to learn the 
signals that are particular to the corresponding input  
without the noise from other inputs. 
The MSEN also makes it easy to extend to new inputs without 
losing the knowledge the model has learned from existing streams.
In this paper, we explore MSEN approaches that could better embed context information for a persona-based dialogue system. 
Note that our method focuses on encoding the context better, therefore
it can be easily extended to generation tasks.}

\newtext{\textbf{Pre-fusion-based methods}~\mbox{\cite{humeau2019poly, chen2020sequential}} typically concatenate the multi-inputs before encoding (e.g., $h = g([P; H; q])$ in~\mbox{\cite{humeau2019poly}}) and implicitly assume an ordering among the inputs, which is not always true (e.g., the persona and query are independent). 
Moreover, these methods only work with homogeneous data (e.g., all inputs are textual) and would fail when inputs are heterogeneous, in terms of both the modality and the information they carry. 
\textbf{Muti-stream embedding networks} (MSEN)~\mbox{\cite{gadiraju2020multimodal}} encodes different inputs in separate streams before fusion. Under MSEN, a model is able to embed multi-modal heterogeneous inputs with custom-engineered embedders, each of which is specialized to learn the signals particular to the corresponding input. 
The MSEN is also easily extendable to new inputs without 
losing the learning from existing streams.
In this paper, we focus on MSEN approaches that could better embed context for a persona-based dialogue system, therefore
they can also be easily extended to generation tasks.}

\section{Method}
\label{sec:method}


\begin{figure*}[t]
\vspace{-10pt}
\centering
\begin{subfigure}{0.48\linewidth}
  \centering
  \includegraphics[width=\textwidth]{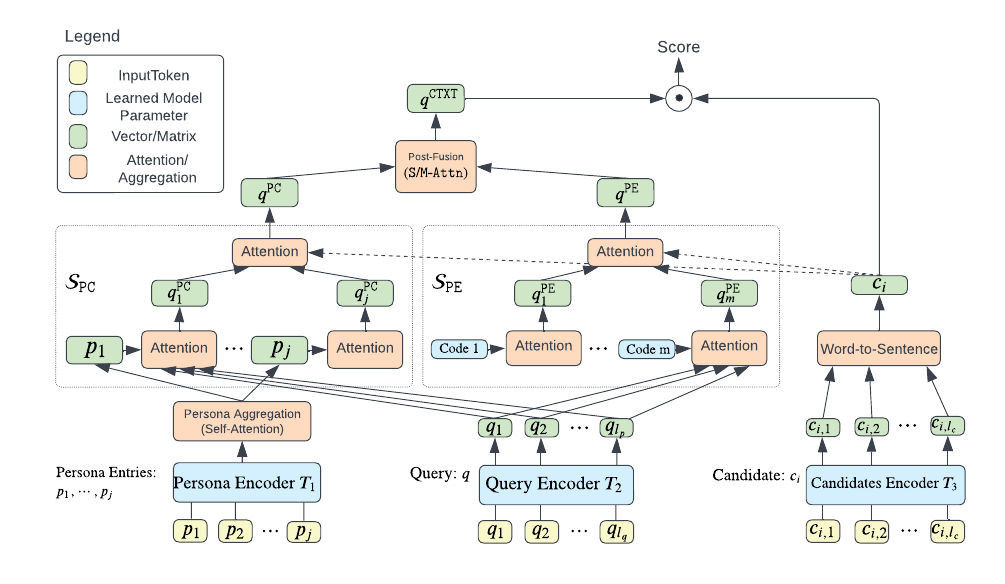}
  \vspace{-20pt}
  \caption{\pcpe with Attention-based Post-Fusion}
  \label{fig:pcpe-attn}
\end{subfigure}
\begin{subfigure}{0.48\linewidth}
  \centering
  \includegraphics[width=\textwidth]{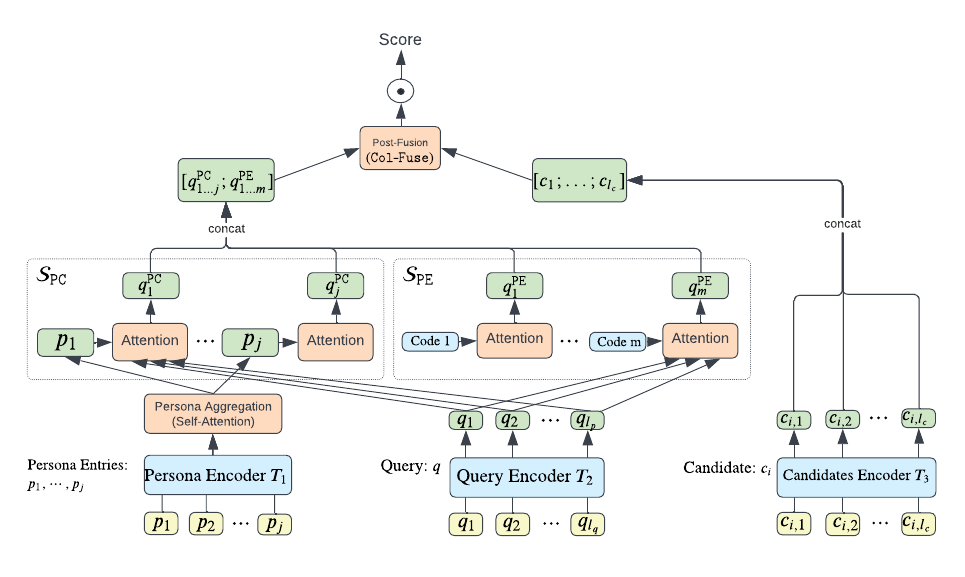}
  \vspace{-20pt}
  \caption{\pcpe with \colbertfusion Post-Fusion}
  \label{fig:pcpe-colfuse}
\end{subfigure}
\vspace{-3pt}
\caption{Network Architecture of \pcpe}
\vspace{-15pt}
\label{fig:network-pcpe}
\end{figure*}

Our method \pcpe, shown in Fig.~\ref{fig:network-pcpe}, 
embeds the conversation context under the MSEN scheme by leveraging
two separate processing streams: 
\edittext{the \mbox{\PEstream} stream that creates a poly-encoded
context embedding in Section~\mbox{\ref{sec:method:pcpe:poly}}, 
and the \mbox{\PCstream} stream that creates a persona-encoded 
context embedding in Section~\mbox{\ref{sec:method:pcpe:persona}} (see also the dotted boxes denoted \mbox{\PEstream} and \mbox{\PCstream} in Fig.~\mbox{\ref{fig:network-pcpe}}).
Fig.~\mbox{\ref{fig:pcpe-attn}} and Fig.~\mbox{\ref{fig:pcpe-colfuse}} show two different version of \mbox{\pcpe}
with different post-fusion methods (see Section~\mbox{\ref{sec:method:pcpe:fusion}}).}
{the persona-encoded context embedding stream (\mbox{\PCstream}) and the poly-encoded context embedding stream (\mbox{\PEstream}).}
\edittext{In the two streams, three}{Three}
transformer encoders,
$T_1$, $T_2$ and $T_3$, are created to embed persona, query and
candidates, respectively, that is, 
$p_j = T_1(\text{persona})$,
$q = T_2(\text{query})$,
$c_i = \mathcal{R}(T_3(\text{cand}))$,
\edittext{where $T(\cdot)$ is the transformers' output, }{where}
$\mathcal{R}(\cdot)$ is a word-to-sentence reduction (e.g., mean),
$p_j \in \real^{l_p\times d}$, 
$q \in \real^{l_q\times d}$, $c_i \in \real^{d}$, 
$l_p$ is the length of the persona entry,
$l_q$ is the length of the query, and
$d$ is the embedding size.
Note that one could share the encoders, i.e., $T_1 = T_2 = T_3$.

The query embedding $q$ and the persona embeddings $p_j$'s interact
with the candidate embeddings in a similar way as \pe~\cite{humeau2019poly}
in the \PEstream and \PCstream streams, respectively.
The output of the two streams are later fused together as the
global context embedding, which is used to generate the similarity 
scores with the candidates for ranking. 
Note that the baseline \pe method uses only the output of the
single \PEstream stream as the global context embedding.


\subsection{Persona-Coded Context Stream}
\label{sec:method:pcpe:persona}
The persona-coded context stream (\mbox{\PCstream}) learns an embedding of 
the persona entries and the query jointly.

\subsubsection{Persona Embeddings}
The transformer $T_1$ outputs low-level embeddings of a persona entry
$p_j \in \real^{l_p\times d}$, which is first aggregated into
embedding $p_j \in \real^{d}$. 
For text-based personas, the aggregation means word-to-sentence reduction for each person entry (i.e., $j$=10 for two speakers),
and for attribute-based personas, it is for all KV pairs of each speaker (i.e., $j$=2). 
Instead of simply using the mean vector of all low-level embeddings, 
here we consider a self-attention layer (\mbox{\selfattention}) for the aggregation, inspired by~\cite{vaswani2017attention}. 
The \selfattention layer learns self-attention weights of the low-level elements and aggregates according to the weights, i.e.,
\begin{equation}
\label{eqn:pcpe:p-reduction}
    p_j =  \sum_i \alpha_i p_{j, i},
\end{equation}
where $p_{j, i} \in \real^d$ is the embedding of the $i$-th element 
(e.g., $i$-th word or KV) in a persona entry $p_j$, weights vector $\boldsymbol{\alpha} \in \real^{l_p}$ is learned as
$(\alpha_1,\cdots, \alpha_{l_p}) = \softmax(q_{j, 1} \cdot w_p, \cdots, q_{j, l_q} \cdot w_p)$, 
and $w_p \in \real^{d\times 1}$ is a trainable projection matrix.

\subsubsection{Query Embeddings}
We use encoder $T_2$ to generate word-level embeddings of the query. Here we do not reduce the query embeddings to sentence level. 

\subsubsection{Persona-Coded Query Embeddings}
For each persona embedding $p_j$, we generate a persona-coded context embedding
$q^{\texttt{PC}}_1, \cdots, q^{\texttt{PC}}_j$
by attending  $q$ over each of the $p_j$'s as
\begin{equation}
\label{eqn:prsn:q_PC_j}
    q^{\texttt{PC}}_j = \sum_{i=1}^{l_q}  w_i^{\texttt{PC}} q_i
\end{equation}
where $q^{\texttt{PC}}_j \in \real^{d}$, the weights $w_i^{\texttt{PC}}$'s are the interaction between $q$ and $p_j$ as
$(w_1^{\texttt{PC}},\cdots, w_{l_q}^{\texttt{PC}}) = 
    \softmax(p_j \cdot q_1, \cdots, p_j \cdot q_{l_q})$.

\subsubsection{Candidate-Aware Persona Context Embeddings}
Then $c_i$ is attended over the $j$ persona-coded context embeddings.
This layer further explores the relevance between the candidate and the context,
particularly the persona entries. The output candidate-aware context embedding
with persona, $q^{\texttt{PC}}$, is used as the output of the \PCstream stream, and is calculated as 
\begin{equation}
\label{eqn:prsn:q_PC}
    q^{\texttt{PC}} = \sum_{i}^{m} w_i q^{\texttt{PC}}_i,
\end{equation}
where $q^{\texttt{PC}} \in \real^{d}$ and the attention weights are calculated
as \\
$(w_1,\cdots, w_j) = 
        \softmax(c_i \cdot q^{\texttt{PC}}_1, \cdots, c_i \cdot q^{\texttt{PC}}_j)$.
        


\subsection{Poly-Encoded Context Stream}
\label{sec:method:pcpe:poly}
The poly-encoded context embedding stream (\PEstream)
inherits the architecture from the baseline \pe. 
It embeds context in a similar way to \PCstream and 
uses the same query embeddings from $T_2$
as in \PCstream. 
The poly-encoded query embeddings $q^{\texttt{PE}}_1, \cdots, 
q^{\texttt{PE}}_m$ and the output $q^{\texttt{PE}}$ of the stream are 
generated similarly as Equation~\ref{eqn:prsn:q_PC_j} and~\ref{eqn:prsn:q_PC}.
The difference is that
in \PEstream, the query embedding $q$ is attended over $m$ 
trainable codes  $K = [k_1; \cdots; k_m] \in \real^{m \times d}$,
instead of the $j$ persona entry embeddings in \PCstream.
Each of $q^{\texttt{PE}}_i$'s corresponds to one of the $m$ codes.
The rest of the \PEstream stream is similar to $q^{\texttt{PC}}$.
%
%

The $m$ coded-context embeddings $q^{\texttt{PE}}_m$ can be viewed as $m$
different points of views (POVs) to understand the input query, which are controlled
by the $m$ codes. As the model is trained, the $m$ codes will also adjust their
way of viewing the query. 
Whereas in \PCstream stream, it uses $j$ POVs which are more 
specific than the $m$ trainable directions as they explicitly seek
the high-level relevance between the persona entries and the query. 
Note that, \pcpe allows $m=0$ (the \PEstream stream is not effective),
while the baseline \pe requires $m>0$.

\subsection{Post-Fusion}
\label{sec:method:pcpe:fusion}
We fuse the outputs of the \PEstream and \PCstream
($q^{\texttt{PE}}$ and $q^{\texttt{PC}}$) to 
create a global context embedding $q^{\texttt{ctxt}}$ w.r.t the candidate. 
We consider two options based on attention mechanisms: 
self-attention (\selfattention) and multi-level attention (\multiattention ),
as in Fig.~\ref{fig:pcpe-attn}. 
\selfattention works the same as Equation~\mbox{\ref{eqn:pcpe:p-reduction}}.
\multiattention attends again over the candidate embedding $c_i$ similar to Equation~\mbox{\ref{eqn:prsn:q_PC}}.
That is,
$
q^{\texttt{ctxt}} = w_1 \cdot q^{\texttt{PC}} + w_2 \cdot q^{\texttt{PE}},
$
where 
\[
(w_1,w_2) = \begin{cases}
    \softmax(w_f \cdot q^{\texttt{PC}}, w_f \cdot q^{\texttt{PE}}), & \text{with ``\selfattention''}, \\   
    \softmax(c_i \cdot q^{\texttt{PC}}, c_i \cdot q^{\texttt{PE}}), & \text{with ``\multiattention'',}
\end{cases}
\]
$w_f$ is a fully trainable weight matrix to the \selfattention layer and $c_i$ is the candidate embedding to the \multiattention layer.

Instead of fusing the high-level final outputs of \PEstream and \PCstream, we consider 
a third fusion (denoted as \colbertfusion, as in Fig.~\ref{fig:pcpe-colfuse}), inspired by ColBERT's low-level contextualized 
late interaction~\cite{khattab2020colbert}. 
With \colbertfusion, we concatenate the intermediate outputs from two streams into 
$\tilde{q} = [q^{\texttt{PC}}_{1...j}; q^{\texttt{PE}}_{1...m}]$, and 
calculate an interaction score matrix $S_{col} = \tilde{q} \cdot c_i (S_{col} \in \real^{(j+m)\times l_c})$ between the low-level signals
in $\tilde{q}$ and $c_i$ (word embeddings prior to aggregation). Then we follow standard ColBERT score calculation, that is,
take the maximum score in $S_{col}$ along the dimension corresponding to words in $c_i$, then
sum along the remaining dimension of ($j$+$m$) scores. 
The final sum represents the similarity score between the context and candidate $c_i$
and will be used later directly for ranking.


\subsection{Ranking}
\label{sec:method:pcpe:ranking}
For attention-based post-fusion methods (\selfattention and \multiattention), 
the ranking score $s_i \in \real$ of the candidate $c_i$ is calculated as
the dot product with the post-fused global context embedding $q^{\texttt{ctxt}}$,
i.e.,
$s_i = \sigma(q^{\texttt{ctxt}} \cdot c_i)$,    
where $\sigma(\cdot)$ is the Sigmoid function.
For the ColBERT-style post-fusion (\colbertfusion), the ColBERT-score is used directly as the ranking score $s_i$.
A response is sampled based on scores $s_i$'s of all candidates in $C$.
The \pcpe model is trained to minimize cross-entropy
loss over the scores/logits of the candidates.
\section{Experiments}
\label{sec:experiments}

\subsection{Dataset}
\label{sec:exp:dataset}
We compare our methods with the baseline methods (introduced in Section~\ref{sec:exp:baseline})
on two benchmark persona-based conversational datasets: 
\textbf{PersuasionForGood} (\textbf{PFG}) dataset~\cite{wang2019persuasion}
and the NeurIPS \textbf{ConvAI2} dataset~\cite{dinan2020second}. 
Both datasets contain conversations consisting of the persona entries of the two speakers, 
a sequence of chat history and a query.

\textbf{PFG}:
The PFG dataset consists of 1017 conversations, along with persona attributes of demographic features and
psychological assessment scores from user surveys. We randomly split 80\%/20\% (813/204 conversations)
for training/validation purposes. 
The two speakers are randomly paired up and assigned ``persuader'' and ``persuadee'' roles, respectively.
The persuader is asked to persuade the persuadee to donate to a charity through psychological strategies. 
The dialogue reflects the assigned personas of the speakers and the persuasion strategies used by the persuader.
For example, 
a persuadee with persona ``age: 35, rational: 4, ...'' ($p_i$)
asked ``... where does that money go towards?'' ($q$),
the persuader with persona  ``age: 30, rational: 5, ...'' ($p_j$)
responded ``The money goes towards providing meals and clean water.'' ($c^*$).

\textbf{ConvAI2}:
The ConvAI2 dataset consists of 19,893 dialogues, among which 17,878 are 
for training and 1,000 are for validation. 
Each speaker is assigned 4$\sim$5 persona entries from a total of 1,155 unique person entries.
Each persona entry is a short sentence description of the speaker.
Then they are randomly paired and asked to get to know each other through dialogue.
The dialogue reflects the assigned personas of the speakers.
For example, if a speaker is assigned  a persona entry
``I am a fan of Michael Jordan'' ($p_i$), when he was asked ``What do you do at leisure time?'' ($q$), he might respond ``I watch a lot of basketball games.'' ($c^*$).

\subsection{Baseline Methods}
\label{sec:exp:baseline}
We consider two strong retrieval-based methods with pre-fusion approaches as our baselines: \pe~\cite{humeau2019poly} and \colbert~\cite{khattab2020colbert}. 
Neither of the methods is designed to specifically handle persona data.
The persona data are pre-fused with the queries as longer text inputs to these two methods. 
For the ConvAI2 dataset, the text descriptions of persona entries can be directly pre-fused with the query. 
For PFG, we convert the KVs as a long string delimited by colons and commas (e.g., ``k1 : v1 , k2 : v2 , ...'')
then pre-fuse it with the query. 

\subsection{Experimental Setup}
\label{sec:exp:setup}
\edittext{Our experiments are implemented with the ParlAI framework (\mbox{\url{https://parl.ai}})
from Facebook AI.
%
We use the existing \mbox{\pe} method and ConvAI2 task implementation, and newly 
implemented the \mbox{\colbert} method, our \mbox{\pcpe} method and PFG task in ParlAI framework using
PyTorch
(Note: following ParlAI's terminology, we use ``task'' and ``dataset'' interchangeably in this paper).}
{Our experiments are conducted using ParlAI framework (\mbox{\url{https://parl.ai}}) with existing \mbox{\pe} method and ConvAI2 task\footnote{We use ``task'' and ``dataset'' interchangeably following ParlAI terminology} implementation. We implemented the \mbox{\colbert} method, our \mbox{\pcpe} method and PFG task separately\footnote{\mbox{\url{https://github.com/jliu-v/persona-chat}}}.}
We use a separate GPT-2 encoder for personas ($T_1$) and share another GPT-2 encoder
for queries/candidates ($T_2$=$T_3$). 
For ConvAI2, the inputs to $T_1$ are trainable word embeddings and segment embeddings
in the persona entries. 
For PFG, the inputs to $T_1$ are the trainable key embeddings, value embeddings,
and speaker embeddings. 
All encoders are initialized with pre-trained poly-encoder weights with the Reddit dataset, 
except $T_1$ is randomly initialized for the PFG task.

All experiments are trained on NVIDIA Titan and 2080-Ti RTX GPUs with 24GB memory
for 10 epochs (ConvAI2) and 100 epochs (PFG). 
The models are evaluated on a validation set every 15 minutes. 
For training, we use all the true responses from the batch as the shared candidate \newtext{set} for better efficiency.
For validation, each input query is 
assigned a separate set of 20 candidates, among which there is only one true response. 
We follow all setups in~\cite{humeau2019poly} except a smaller training batch size of 32 to
avoid running out of memory. However, the batch size is fixed for all experiments for
fair comparison.  
\edittext{The \mbox{\colbert} models are tuned with hidden embedding size $d_h$ of $[64, 128, 256]$. 
The \mbox{\pe} models are tuned with $m$ of $[5, 16, 64]$.
The \mbox{\pcpe} models are tuned with $m$ of $[0, 5, 16, 64]$ and 
multi-stream post-fusion of $[\selfattention, \multiattention, \colbertfusion]$.}
{We tune different hidden embedding sizes $d_h$ for \mbox{\colbert}, different $m$ values for \mbox{\pe}, 
and different $m$ values and multi-stream post-fusion strategies for \mbox{\pcpe}.}

%

\subsection{Evaluation Metrics}
\label{sec:exp:metrics}
We evaluate our method with common ranking/retrieval metrics that are widely used in 
information retrieval applications.
We follow the work of Liu~\etal~\cite{liu2022persona}
and measure the 
\textbf{hit rate} at top-$K$ ($\hrk{k}$),
\textbf{mean reciprocal rank} ($\mrr$),
\textbf{F-1 score} ($\fscore$), and 
\textbf{BiLingual Evaluation Understudy score} ($\bleun{4}$ based on 4-gram).
All four metrics  range between 0 and 1, and the larger the values are, the better
the model is able to prioritize the true response among all candidate responses. 
Note that $\hrk{1}$ is equivalent to the prediction accuracy.

\section{Experimental Results}
\label{sec:exp:results}
\begin{table*}
\vspace{-10pt}
\caption{Performance Comparison}
\vspace{-10pt}
\begin{center}
\begin{threeparttable}
\bgroup
\def\arraystretch{1}%
  \begin{tabular}{
      @{\hspace{0pt}}c@{\hspace{10pt}}
      @{\hspace{0pt}}c@{\hspace{5pt}}
      @{\hspace{5pt}}r@{\hspace{15pt}}
      @{\hspace{5pt}}c@{\hspace{5pt}}
      @{\hspace{5pt}}c@{\hspace{5pt}}
      @{\hspace{5pt}}c@{\hspace{5pt}}
      @{\hspace{5pt}}c@{\hspace{5pt}}
      @{\hspace{5pt}}c@{\hspace{15pt}}
      @{\hspace{15pt}}c@{\hspace{5pt}}
      @{\hspace{5pt}}c@{\hspace{5pt}}
      @{\hspace{5pt}}c@{\hspace{5pt}}
      @{\hspace{5pt}}c@{\hspace{5pt}}
      @{\hspace{5pt}}c@{\hspace{5pt}}
    }
\hline 
\multirow{2}{*}{Method} & \multirow{2}{*}{Post-Fusion} &   & \multicolumn{5}{c}{PFG Dataset}& \multicolumn{5}{c}{ConvAI2 Dataset}\\

 &  &   & $\hrk{1}$ &        $\hrk{5}$ &        $\fscore$ &           $\mrr$ & \bleun{4}
&        $\hrk{1}$ &        $\hrk{5}$ &        $\fscore$ &           $\mrr$ & \bleun{4}\\
\hline
\multirow{12}{*}{\mbox{\pcpe}} 
  &  \multirow{4}{*}{\selfattention}
    &      $m=0$ &      0.849 &      0.982 &      0.866 &      0.907 &      0.849 & \bf{0.666} &      0.932 & \bf{0.707} &      0.777 & \bf{0.623}        \\
  & &      $m=5$ &      0.852 &      0.983 &      0.869 &      0.909 &      0.852 &      0.652 &      0.931 &      0.694 &      0.771 &      0.605        \\
  & &     $m=16$ &      0.851 &      0.983 &      0.868 &      0.909 &      0.851 &      0.646 &      0.931 &      0.690 &      0.770 &      0.603        \\
  & &     $m=64$ &      0.851 &      0.984 &      0.868 &      0.909 &      0.851 &      0.648 &      0.926 &      0.692 &      0.768 &      0.606        \\

    \cline{2-13}
    
  & \multirow{4}{*}{\multiattention}
    &      $m=0$ &      0.848 &      0.981 &      0.866 &      0.907 &      0.848 & \bf{0.666} & \bf{0.941} &      0.706 & \bf{0.782} &      0.621        \\
  & &      $m=5$ &      0.845 &      0.982 &      0.862 &      0.905 &      0.844 &      0.655 &      0.930 &      0.696 &      0.774 &      0.612        \\
  & &     $m=16$ &      0.844 &      0.982 &      0.862 &      0.904 &      0.844 &      0.627 &      0.914 &      0.673 &      0.753 &      0.586        \\
  & &     $m=64$ &      0.851 &      0.981 &      0.867 &      0.908 &      0.851 &      0.653 &      0.926 &      0.695 &      0.773 &      0.610        \\
    
    \cline{2-13}
    
  & \multirow{4}{*}{\colbertfusion}
    &      $m=0$ &      0.852 &      0.982 &      0.869 &      0.910 &      0.852 &      0.628 &      0.916 &      0.671 &      0.751 &      0.585        \\
  & &      $m=5$ &      0.847 &      0.982 &      0.864 &      0.906 &      0.847 &      0.621 &      0.921 &      0.666 &      0.750 &      0.579        \\
  & &     $m=16$ &      0.846 &      0.982 &      0.863 &      0.905 &      0.846 &      0.616 &      0.915 &      0.662 &      0.741 &      0.570        \\
  & &     $m=64$ &      0.705 &      0.944 &      0.739 &      0.807 &      0.705 &      0.401 &      0.780 &      0.467 &      0.567 &      0.370        \\

\hline
\multirow{3}{*}{\mbox{\colbert}}
  & \multirow{3}{*}{-} 
    &   $h_d=64$ &      0.850 & \bf{0.986} &      0.866 &      0.909 &      0.850 &      0.603 &      0.921 &      0.650 &      0.738 &      0.562        \\
  & &  $h_d=128$ & \bf{0.855} &      0.985 & \bf{0.871} & \bf{0.912} & \bf{0.855} &      0.574 &      0.929 &      0.633 &      0.725 &      0.529        \\
  & &  $h_d=256$ &      0.852 & \bf{0.986} &      0.868 &      0.911 &      0.852 &      0.647 &      0.933 &      0.689 &      0.770 &      0.603        \\

\hline
\multirow{3}{*}{\mbox{\pe}}  
  & \multirow{3}{*}{-} 
    &      $m=5$ &      0.837 &      0.981 &      0.856 &      0.900 &      0.837 &      0.644 &      0.929 &      0.689 &      0.768 &      0.599        \\
  & &     $m=16$ &      0.849 &      0.983 &      0.866 &      0.907 &      0.849 &      0.615 &      0.929 &      0.662 &      0.749 &      0.572        \\
  & &     $m=64$ &      0.853 &      0.984 &      0.869 &      0.911 &      0.853 &      0.647 &      0.923 &      0.689 &      0.765 &      0.599        \\

\hline
\hline
\end{tabular}
\begin{tablenotes}[flushleft]
    \setlength\labelsep{0pt}
    \footnotesize
    \item Values in \textbf{bold} represent the best performance of the corresponding metric among all methods.
\end{tablenotes}
\egroup
\end{threeparttable}
\end{center}

\vspace{-20pt}
\label{tbl:performance}
\end{table*}


Our experimental results showed that \pcpe is able to effectively utilize 
persona information in response selection. 
On the PFG task with multi-modal persona data,
multi-stream-based \pcpe is able to significantly improve the conversation quality over 
the state-of-the-art single-stream pre-fusion-based methods.
On the ConvAI2 task with text-based persona data, \pcpe is able to maintain similar performance
as both of the baseline methods. 
Table~\ref{tbl:performance} compares the performance of \pcpe with the baseline \pe and \colbert models. 
%
\edittext{We do not report the results of other experimental setups, such as the margin ranking 
loss function, since they did not show very competitive outcomes. 
However, we will give a brief discussion in Section~\mbox{\ref{sec:exp:results:suboptimal}}.}
{We will discuss in Section~\mbox{\ref{sec:exp:results:suboptimal}} other sub-optimal setups we eliminated from the report.}

\subsection{Overall Performance}
\label{sec:exp:results:overall}
On the PFG task, the best performance of \pcpe is able to surpass that of the best baseline (\colbert) by 
2.94\% (0.666 vs. 0.647) in terms of accuracy/\hrk{1},
2.61\% (0.707 vs. 0.689) in terms of \fscore,
0.91\% (0.777 vs. 0.770) in terms of \mrr, and
3.32\% (0.623 vs. 0.603) in terms of \bleun{4}. 
With different attention-based post-fusion aggregation methods, \pcpe is able to 
outperform \pe in most cases. 
\pcpe achieved \hrk{1} = 0.666 when using \selfattention/\multiattention,
while the same best metric for the baselines is 0.647. 
These results validate our primary hypothesis that embeddings over simple concatenated multi-modal data inputs do not fully capture the rich information and relationships among the attributes. Whereas our \pcpe method, by employing attribute-specific streams to capture specialized information, and post-fusion to learn multi-modal relationships overcomes the limitations of existing pre-concatenated single-stream embeddings.

On the ConvAI2 task, our model is able to maintain similar performance with the best baseline. 
The difference between the \pcpe and the best \colbert model is only 
0.35\% (\hrk{1} 0.852 vs. 0.855),
0.23\% (\fscore 0.869 vs. 0.871),
0.33\% (\mrr 0.909 vs. 0.912), and
0.35\% (\bleun{4} 0.852 vs. 0.855).
We anticipated this result since the persona data in the ConvAI2 task is still text-based,
and \pcpe is expected to work better with multi-modal inputs. 
We also inspected the query-response pairs where the best \pcpe made mistakes 
and the best baseline \colbert was correct. Many of the \pcpe's ``mistakes'' 
surprisingly made sense, 
such as in greeting types of conversations where ``I am good'' and ``I am doing alright'' are both acceptable when the speaking
partner asks ``how are you?''.
In these cases, \pcpe made good selections although
another candidate was labeled true by the dataset.
\textbf{Therefore, we anticipate that the actual performance of \pcpe should be higher than
\colbert after this adjustment}.
We will also discuss this challenge in Section~\ref{sec:discussion:challenges}.

\subsection{Effectiveness of $m$}
\label{sec:exp:results:m}
With the \pcpe framework, we seek effective ways to encode persona information. 
Recall in Section~\ref{sec:method:pcpe:poly}, $m$ and $j$ can be 
viewed as different POVs to understand the query. 
$j$ is the POVs directed by the persona, which is fixed by the input data. 
In this part, we study how $m$ affects the way that the \mbox{\pcpe}
model encodes the persona entries.

On the PFG task, \pcpe achieved its best performance
when $\mathbf{m=0}$ and only the \PCstream
stream is effective. The POVs are fully dictated by the personas (ref. Section~\ref{sec:method:pcpe:fusion}). 
This indicates that replacing the $m$ randomly trainable codes from \pe with actual 
persona information is helpful for the model to learn better. In addition, this empirically
proved our claim in Section~\ref{sec:method:pcpe:poly} that the persona-coded directions are
more specific than the $m$ random directions for the model to understand the relevance between
the context and response and guide the response selection. 
When $\mathbf{m \ge 0}$ the performance of all methods decreases in general when $m$ increases,
but the \pcpe is still able to outperform the baselines in most cases. 
This is mainly due to the fact that the PFG dataset is goal-specific (for persuasion purpose)
and the conversations have higher correlation with the persona data, especially the psychological
and behavioral attributes. The randomly trained POVs (when $m>0$) might bring in extra 
noise in the model, which could be confusing and detrimental to the learning.

On the ConvAI2 task, \pcpe achieved the best performance with small $m$ values 
($m=0$ for \multiattention/\colbertfusion and $m=5$ for \selfattention fusion). 
This is in general consistent with the conclusion with the PFG task. 
However, as $m$ increases, the performance doesn't necessarily decrease. 
This is probably because the ConvAI2 dataset is a chit chat dataset without a specific purpose,
and the correlation between the persona and conversation is not as strong as in the PFG dataset. 
Therefore, a non-zero $m$ is able to supplement
information in case the persona entries are not able to provide sufficient context
(e.g., not all utterances are about persona, for example, some generic greeting conversations),
but excessively large $m$ values could still be detrimental and add complexity to the model.

\subsection{Effectiveness of Post-Fusion Methods}
\label{sec:exp:results:fusion}
In our experiments, we compare two attention-based post-fusion methods 
(\mbox{\selfattention} and \mbox{\multiattention}) and 
the ColBERT-style fusion (\colbertfusion).
From Table~\ref{tbl:performance}, \selfattention and \multiattention post-fusions 
achieved similar and the best overall performance. This \edittext{might be}{is} attributed to the non-linearity introduced by
the attention mechanism. 
\edittext{The performance of \mbox{\selfattention} is slightly better than \mbox{\multiattention}
in some cases, but they are very close to each other.
This is probably because the information from candidate embedding has already been
efficiently extracted in the \mbox{\PCstream} stream, and exploiting twice in the post-fusion
stage with \mbox{\multiattention} is not able to extract additional information that's helpful
for the model to get significant improvement over \mbox{\selfattention}.}
{The performance of \mbox{\selfattention} is only marginally better than \mbox{\multiattention}.
This is probably because the \mbox{\PCstream} stream has already sufficiently extracted information 
from the candidate, and exploiting it twice during post-fusion is not able to generated additional useful signal
to improve the performance.}

Our experiments also showed that \colbertfusion doesn't have as good performance as the attention-based post fusions. 
Unlike attention-based post-fusions where the stream outputs were aggregated, the \colbertfusion scoring considers all low-level inputs ($q^{\texttt{PC}}_1, \cdots, q^{\texttt{PE}}_m$) equally with the final sum.
Thus, irrelevant inputs (that are not contributing to the conversation) could dominate the 
scoring function and thus yield sub-optimal results.
This can also be confirmed by the fact that larger $m$ leads to lower performance when using \colbertfusion fusion
as more irrelevant inputs can be introduced.

\subsection{Other Sub-optimal Setups}
\label{sec:exp:results:suboptimal}

\edittext{We now briefly present a few additional choices that we explored. Though the results were sub-optimal, we hope they offer additional insights and may provide paths for further explorations.}
{We now briefly present a few sub-optimal choices we explored, and hope they offer additional insights and may provide paths for further studies.}

\paragraph{Persona/Candidate Aggregation}
Linear aggregation methods (mean and sum) didn't outperform
\selfattention because linearly combing word/KV embeddings might dilute the 
signals carried by certain key words/attributes, especially the ones that are
relevant to the conversation. 
\paragraph{Post-Fusion Methods}
\edittext{We also considered other linear options for post-fusion, including sum and concatenation.}
{We consider linear post-fusion options (e.g., sum and concatenation).}
These fusion methods treat all streams (or all inputs) as equally important to the 
task, which is not often true in many different application domains. 
%
%
\paragraph{Loss Functions}
We also tried margin ranking loss~\footnote{\url{https://pytorch.org/docs/stable/nn.html\#loss-functions}}
to train the model by arranging the ranking list into
true-false response pairs and optimize the score margin. 
This ranking loss didn't result in better performance, 
probably because, with only one candidate is correct, 
the candidates set lacks necessary preference ordering
among all or most of the candidates, which is required
by many other ranking problems.

\section{Discussions}
\label{sec:discussion}

\subsection{Multi-Modal Processing Stream}
\label{sec:discussion:multimodal}
As stated earlier, the main purpose of this paper is to explore ways to better utilize heterogeneous auxiliary information in different modalities to improve agent conversations. 
Although \pe and \colbert are strong retrieval models, they are unable to effectively utilize multi-modal data as it has only one stream that processes a single text input.

We conducted additional experiments of \pcpe on the PFG task where the personas are treated as 
long text inputs (denoted as \pcpetext) in a similar way as the ConvAI2 task with $T_1$ for text,
instead of KV attributes (the regular \pcpe, denoted as \pcpekv) with a different $T_1$ for KV.
The text-based personas resulted in the best \hrk{1} of 0.654, compared to the best \hrk{1} of 
0.647/0.666 of \colbert/\pcpekv. 
This not only demonstrated superiority of the post-fusion over the
single-stream pre-fusion (\pcpetext outperforms \colbert), 
but it also showed the necessity of using multi-modal processing streams
(\pcpekv outperforms \pcpetext),
especially when the auxiliary inputs are naturally heterogeneous.


Moreover, the multi-modal stream framework provides new opportunities to effectively
utilize various auxiliary information that might be specific to a domain or an application.
For example, for Precision Nudging, one could easily 
incorporate the patients' health records, behavior patterns, etc. in new streams without
re-training the entire conversation model.


\subsection{Computational Efficiency}
\label{sec:discussion:computation}
At inference time, \pcpe has several computational advantages. 
Like \pe and \colbert, the candidate embeddings can be pre-computed and cached. 
This is especially beneficial when the number of candidate responses is large.
In addition, the persona embeddings can also be cached offline with \pcpe, 
while it is not achievable with baseline single-stream pre-fusion-based methods
as the queries are unknown prior to inference time. 
The storage cost for the persona embeddings is linear with the number of speakers $O(N)$,
given that the number of persona entries $j$ is fixed for each speaker.

At training time, we note that \pcpe introduced more training parameters and therefore may require longer training. But this could be remedied by using pre-trained 
encoders and fine-tuning, sharing the transformer encoders in different
streams, etc. Empirically from our experiments, \pcpe needs a smaller $m$ value to get the best 
performance ($m=0$ or 5). This made \pcpe more efficient than \pe
at both training and inference time.
\colbert calculates the word-level interaction between the context and candidates,
and thus is much more inefficient than \pcpe and \pe.

\subsection{Challenges and opportunities}
\label{sec:discussion:challenges}

\subsubsection{Lack of a Realistic Persona Dataset}
One major challenge for personalizing conversation is that there is no
publicly available dataset that provides easily accessible persona
data for large-scale applications. 
The cost and privacy issues with text-based personas, like in ConvAI2,
remain a concern for real-world applications.
Although the PFG dataset provides feature-based personas, it is still not ideal
for exploring other forms of personas that are helpful for personalization in
various applications, such as user health histories.
%
%
As far as we know, other public datasets, such as Reddit~\cite{mazare2018training}
and Twitter~\cite{ritter2010unsupervised}, are not ideal for personalization since 
they contain only the user identifiers without actual persona information.

\subsubsection{Lack of a Behavior-Driven Dataset}
Another challenge in personalization for conversational AI is the lack of
strong behavior-driven strategies in available datasets that allow one to
navigate the speakers' barriers or motivations, which is needed to provide
the assistance the users truly need.
The PFG dataset is a great example that incorporates persuasion strategies
into chats and influences others' behaviors. However, 
the persuaders have no visibility to the persuadees' personas. 
In addition, the persuaders, as crowd-source
workers, are not professionally trained behavioral scientists.
It is hard to guarantee that effective behavioral strategies are used in the conversations
and personalized to the persuadees.
%
Additional collaboration and guidance from domain experts in behavioral science 
is necessary for the process of generating a rich behavior-driven and persona-based
conversational dataset. 

\subsubsection{Lack of Evaluation on Personalization}
The lack of evaluation metrics for personalization is another main challenge
of persona-based conversations. 
Common evaluation metrics for retrieval methods, such as \hr, do not really consider
the actual contents and only treat the responses atomically as a binary result (true 
or false responses). 
For generation-based methods, evaluations (e.g., BLEU~\cite{papineni2002bleu} and 
Perplexity~\cite{hofmann2013probabilistic}) usually focus
only on the word overlap with a reference response. 
They fail to address the relevance of the response to the speakers'
persona information.
Currently, human expert evaluation is still the dominant way to evaluate
personalization. Such extensive human efforts and domain expertise make it
hard to scale in real applications. Further research efforts are 
needed to explore computational methods that efficiently evaluate the personalization 
of dialogue systems.

\subsubsection{Legal and Ethical Challenges}
Other challenges in persona-based conversational AI research are related
to legal and ethical issues. Persona information can potentially include PHI and PII, which legally require certain standards be observed in their use. Other concerns center around ensuring that a 
conversational agent is not discriminatory or offensive.
\edittext{Although these concerns do not fall within the scope of this paper, we}
{These concerns not only affect the legality but also the reasonableness of a fair, inclusive, justifiable and effective conversational agent. We} encourage the reader to refer to studies that focus on AI ethics for 
personalization~\mbox{\cite{Hermann-22,Libai-20}}.

\section{Conclusion}
\label{sec:conclusion}

In this paper, we presented a novel multi-stream Persona-Coded Poly-Encoder network that admits heterogeneous auxiliary information and uses various post-fusion strategies to obtain a latent representation.
We evaluated the PCPE and compared its performance against the SOTA methods on two benchmark persona-based conversation datasets.
In addition to providing a review of the SOTA in conversational AI for personalization, we identified several limitations and provided insights for future research.

We note that the experimental results demonstrated that our method \pcpe outperforms the baseline \pe and \colbert for response selection tasks with multi-modal inputs.
Our method also illustrates opportunities for leveraging auxiliary multi-modal data in conversational models to further improve the quality of conversations.  However, we have also observed several limitations of current datasets and evaluation metrics. We suggested future research directions to address several key limitations of existing research on persona-based conversations, including the lack of realistic persona and behavior-driven conversational data and the lack of evaluation metrics on personalization.

\bibliographystyle{IEEEtran}
\bibliography{references}

\end{document}


\title{\Large Persona-Coded Poly-Encoder\\
\textit{Supplementary Materials}}
\author{
Junfeng Liu\footnotemark[2]\ \footnotemark[4]
\and 
Christopher Symons\footnotemark[3]
\and 
Ranga Raju Vatsavai\footnotemark[3]\ \footnotemark[4]
}


\date{}

\maketitle

\renewcommand{\thefootnote}{\fnsymbol{footnote}}

\footnotetext[2]{Lirio AI Research, Lirio LLC, Knoxville, TN, USA. (jliu@lirio.com)}
\footnotetext[3]{Behavior Reinforcement Learning Lab, Lirio LLC, Knoxville, TN, USA. (csymons@lirio.com, rvatsavai@lirio.com)}
\footnotetext[4]{Dept. Computer Science, North Carolina State University, Raleigh, NC, USA.}
\renewcommand{\thefootnote}{\arabic{footnote}}






\makeatletter
\renewcommand \thesection{S\@arabic\c@section}
\renewcommand\thetable{S\@arabic\c@table}
\renewcommand \thefigure{S\@arabic\c@figure}
\makeatother

\section{Full Experimental Results on \pcpe}
\vspace{-10pt}
\input{tables/tbl_pcpe_perf_full}



